# A Seft-adaptive Multicellular GEP Algorithm Based On Fuzzy Control For Function Optimization

DENG Chuyan[1], PENG Yuzhong[1,2], LI Hongya[1], GONG Daoqing[1], ZHANG Hao[2], LIU Zhiping[3]

(1. School of Computer & Information Engineering, GuangXi Teachers Education University, Nanning 530299, China; 2. College of computer science & technology, Fudan University, Shanghai 200433, China; 3. Zhuang Autonomous Region, Institute of Meteorological Disaster Reduction, Nanning 530299, China)

**Abstract**：To improve the global optimization ability of traditional GEP algorithm, a Multicellular gene expression programming algorithm based on fuzzy control (Multicellular GEP Algorithm Based On Fuzzy Control, MGEP-FC) is proposed. The MGEP-FC algorithm describes the size of cross rate, mutation rate and real number mutation rate by constructing fuzzy membership function. According to the concentration and dispersion of individual fitness values in population, the crossover rate, mutation rate and real number set mutation rate of genetic operation are dynamically adjusted. In order to make the diversity of the population continue in the iterative process, a new genetic operation scheme is designed, which combines the new individuals with the parent population to build a temporary population, and the diversity of the temporary and subpopulation are optimized. The results of 12 Benchmark optimization experiments show that the MGEP-FC algorithm has been greatly improved in stability, global convergence and optimization speed.

**Key words**：gene expression programming; function optimization;fuzzy control; evolutionary algorithm; seft-adaptive algorithm

## 0　引言

函数优化因其极具应用价值又难于求解，一直是热门的研究问题。近年来，大量的研究表明，仿生算法具有优秀的随机搜索性能，非常适合函数优化问题的求解[1-7]。现有仿生算法能在一定程度上解决函数优化问题，但依旧存在很大的改进空间。例如，算法寻优的收敛代数、解的精度、针对不同问题的自适应能力以及全局最优解的搜索等。

基因表达式编程算法（Gene Expression Programming，GEP）是一个被应用于解决优化问题和知识发现的新型智能算法。因其编码简单，解决复杂问题效率高等优点，在各领域得到普遍认可的同时也激发了学者们的研究热情[8-11]。如文献[12]中，Ferreira 初次将 Hzero 算法、GEP-PO 算法用于函数优化并获得令人满意的成果。针对传统 GEP 个体编码的不足，连续编码的 GEP[13]、多细胞 GEP[14]等改进算法被提出。已有研究工作表明，GEP 很适合用于解决函数优化问题，并且多细胞 GEP 编码比基础 GEP 更为灵活,具有更广的搜索空间和更高的解精度。但是，传统 GEP 和多细胞 GEP 在整个迭代过程中均采用固定初始参数和单一遗传操作顺序，随着演化过程的进行，因种群多样性流失，算法容易陷入局部最优解。

本文提出一种模糊自适应的多细胞 GEP 算法（MGEP-FC），通过设计三个模糊控制器，结合模糊数学思想对种群多样性、交叉率、变异率和实数集变异率的大小进行描述和动态调整。另外，设计了新的遗传操作方案，使临时种群的个体数量及种群多样性得以保持和延续。通过十二个复杂函数寻优实验验证了该算法的优秀性能。

## 1　MGEP-FC 算法

### 1.1　基本思想

在进化算法中，子代种群由临时种群通过选择操作获得，而子代种群的多样性决定算法是否容

易陷入局部最优。所以临时种群和子代种群的多样性越优,算法越不容易陷入局部最优解。传统 GEP 算法通过"先交叉后变异"的遗传操作方式生成临时种群,该方式可能导致部分交叉操作生成的优秀个体经过变异操作后被破坏,影响算法的收敛效率[15]。MGEP-FC 算法将遗传操作的顺序和操作方式进行了调整,种群的多样性得到延续。

进化算法的交叉率很大程度上会影响算法的收敛效率,变异率决定算法是否能跳出局部最优找到全局最优解[15]。GEP 在迭代过程中,优秀基因逐渐扩散、种群逐渐收敛的同时也伴随着种群多样性的流失,导致迭代后期算法容易陷入局部最优,此时适当增大算法的变异率以增强种群多样性,以利于算法跳出局部最优。种群多样性较优时,增大算法的交叉率、降低变异率以加快算法收敛。传统 GEP 算法使用固定的变异率和交叉率,无法满足迭代过程中算法对变异率和交叉率的动态需求。

模糊控制(Fuzzy Control)是一种基于模糊数学的控制方法,通过构建控制对象隶属函数和一系列的模糊规则为被控对象提供较合理的、非线性的动态调整方案。模糊控制善于处理控制因素复杂的问题。在实际的研究工作中,结合模糊控制后的系统运行更加高效和稳定[16-18]。因此,MGEP-FC 算法根据种群中个体适应度值的集中和分散程度,通过模糊控制自适应的调节每次迭代的交叉率和变异率,使算法的迭代过程更加高效。

总的来说,本文提出的 MGEP-FC 算法基本思想示意图如图 1 所示。

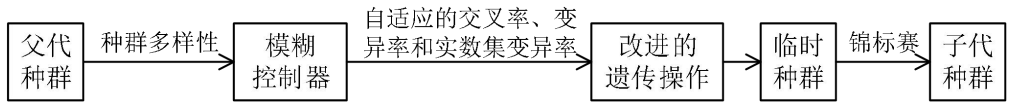

图 1 MGEP-FC 基本思想示意图

### 1.1.1 个体编码

MGEP-FC 算法采用多细胞结构对染色体进行编码,一个个体由若干个普通基因和同源基因组成,一个同源基因表示函数的一个变量值。普通基因和同源基因均包含头部和尾部,除此之外,普通基因还包含 DC(domain constants)域。头部基因可由函数符集和终结符集中的元素组成,尾部只能由终结符集中元素组成,尾部长度必须满足式(1),式中 $h$ 为基因头部长度,$M$ 为函数符最大操目数。

$$t = h \times (M-1) + 1 \tag{1}$$

在 MGEP-FC 中,DC 域中的字母与随机生成的实数集中的元素相对应,实数集生成范围与优化函数的定义域相同,实数集中的元素根据迭代过程中自适应的实数集变异率进行更新;终结符集为{?, a, b, c, d, e, f, g, h, i, j},小写字母 a 到 j 依次表示数字 0 到 9;函数符集为{+, -, *, /, 目标函数专用运算符},其中目标函数专用运算符是指根据待优化的目标函数自身包含的除四则基本运算外的运算表示符号。如,在对本文第 2 节表 4 的 $f_3$ 函数进行最优化时,根据 $f_3$ 包含了正弦函数和平方根函数,而增加目标函数专用运算符正弦函数(简记为 S)和平方根函数(简记为 Q)。在函数集中增加了目标函数专用运算符,实质上是根据待求解问题的实际情况在算法执行过程中加入了领域经验(专家)知识,后文 2.1 节的实验验证了这一操作可提高算法的寻优能力。丰富的函数符集和终结符集及实数基因元素,有助于提高种群的多样性和寻优精度。

图 2 为一个完整个体的基因型,该个体包含两个普通基因和一个同源基因,头长均为 3。普通基因 0 中的元素"?"对应 DC 域的"D",代表实数集中第四个元素 0.87227。图 2 的基因型可得到如图 3 所示的表现型表达式树。通过解码,普通基因 0 表示的数值为 2.47408,普通基因 1 表示的数值为 0.84147,该个体确定的函数变量值为同源染色体映射的值,即 1.49125。如果函数中含有多个变量,则个体包含与之相应个数的同源基因。

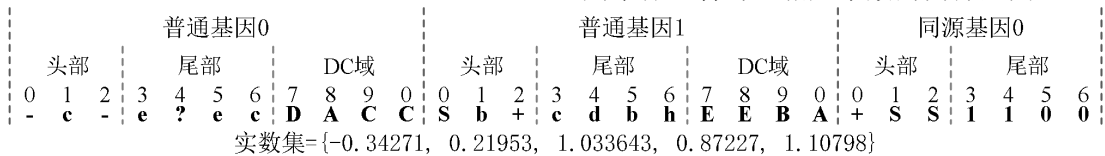

图 2 个体编码示意图

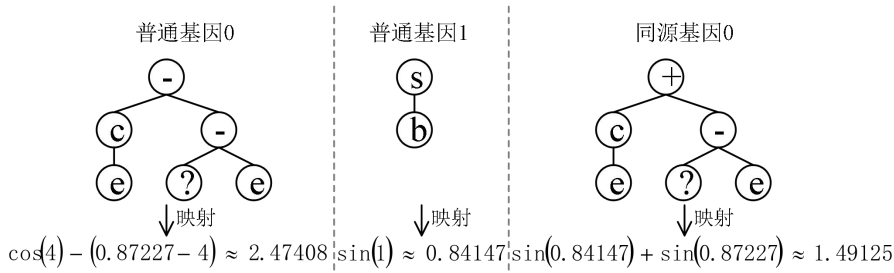

图 3 表达式树

### 1.1.2 种群多样性衡量

演化算法的种群多样性有多种衡量方式[19]。在函数优化问题中，个体适应度值为被优化函数的解，对于不同的优化函数，可能最优解相差甚远，其方差、标准差数值存在一定波动，需要根据不同的函数进行调整，影响算法的鲁棒性。MGEP-FC 通过种群中个体适应度值的集中和分散程度衡量种群多样性，即计算当前种群最优适应度 $F_{best}$ 和平均适应度 $F_{ave}$ 的比值 $d$。求优化函数极小值时，用式(2)确定种群的多样性。反之，则用式(3)进行计算。随着种群收敛，$d$ 逐渐接近 1。根据多次对比实验，选取以下参数方案：当 $d \in [0.6,1]$ 时，把 $d$ 作为模糊控制的输入，自适应调节遗传操作的交叉率和变异率。当 $d<0.6$ 时，为了加快种群收敛，种群的交叉率和变异率分别取固定值 0.4 和 0.1。

$$d = \frac{F_{\min}}{F_{ave}}, F_{best} < F_{ave} \quad (2)$$

$$d = \frac{F_{ave}}{F_{\max}}, F_{best} > F_{ave} \quad (3)$$

### 1.1.3 模糊控制

模糊隶属函数可以计算对象属于某一类别的程度。本文算法通过构建模糊隶属函数分别对种群多样性（范围为[0.6,1.0]）和算法的交叉率（范围为[0.1,0.3]）、变异率（范围为[0.05,0.25]）、实数集变异率（范围为[0.0,0.5]）以及迭代次数（范围为[500,1000]）进行描述。

另外，为交叉率、变异率和实数集变异率分别设计三个不同的模糊控制器，其中控制交叉率和变异率的两个模糊控制器均选取当前种群多样性作为输入，控制实数集变异率的模糊控制器选取当前种群多样性和当前迭代次数作为输入。三个模糊控制器的输出均为下一代种群的交叉率、变异率和实数集变异率。五个模糊隶属函数均选取{较低，中低，中等，中高，较高}五个模糊语言变量进行描述，英文表示为{XL, ML, M, MH, XH}，输入输出隶属函数使用三角隶属函数和梯形隶属函数构建，如图 4 所示。当种群多样性低时，增大算法的变异率以增强种群的多样性，使算法能跳出局部最优。当种群多样性较优时，增大算法的交叉率、降低变异率以加快算法的收敛。根据这一思想，总结出模糊规则表如表 1、表 2、表 3 所示。

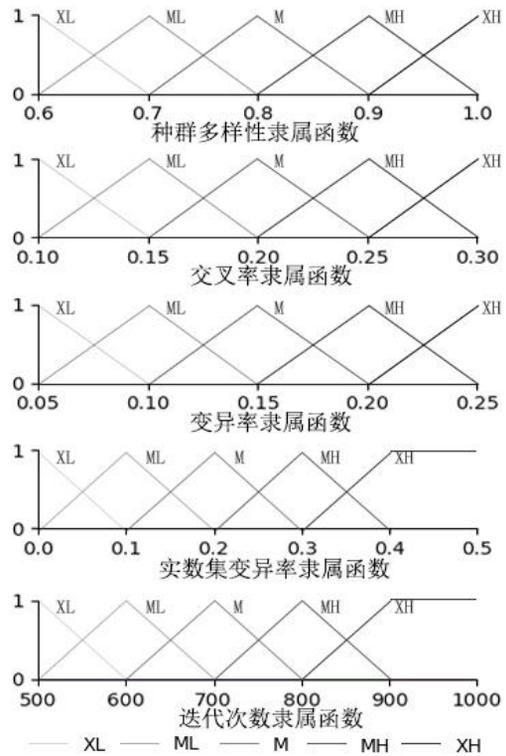

图 4 输入输出隶属函数

表 1 种群多样性与算法交叉率的模糊控制规则表

|  | 种群多样性 | | | | |
|---|---|---|---|---|---|
| 交叉率 | XL | ML | M | MH | XH |
|  | XH | MH | M | ML | XL |

表 2 种群多样性与算法变异率的模糊控制规则表

|  | 种群多样性 | | | | |
|---|---|---|---|---|---|
| 变异率 | XL | ML | M | MH | XH |
|  | XL | ML | M | MH | XH |

表 3 种群多样性与算法实数集变异率的模糊控制规则表

|  | 种群多样性 | | | | |
|---|---|---|---|---|---|
| 迭代次数 | XL | ML | M | MH | XH |
| XL | XL | XL | ML | ML | M |
| ML | XL | ML | ML | M | MH |
| M | ML | ML | M | MH | MH |
| MH | ML | M | MH | MH | XH |
| XH | M | M | MH | XH | XH |

#### 1.1.4 遗传操作

交叉操作是最易产生新个体的遗传操作，但这一操作令基因序列大幅度改变，部分优秀个体基因遭到破坏，并且这一改变是不可逆的。MGEP-FC算法的遗传操作不再使用单一的先交叉后变异操作，而是对父代种群分别进行只交叉、只变异、先交叉后变异、先变异后交叉四种操作，再将这四种遗传操作方案得到的新个体和父代种群的全部个体添加到临时种群。此时，临时种群的多样性提高，各个个体发散性寻优能力更强。改进的遗传操作流程如图5所示。

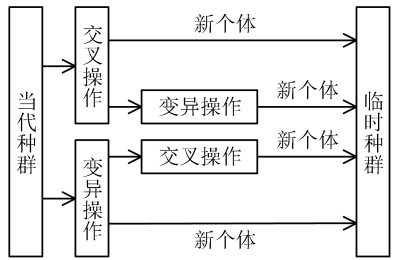

图 5 遗传操作流程图

### 1.2 MGEP-FC 算法流程

根据上述的算法思想，MGEP-FC 算法流程如图 6 所示。

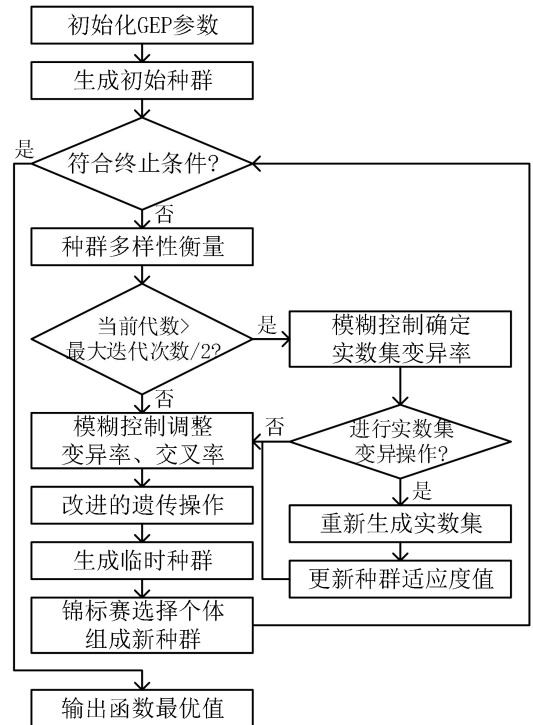

图 6 MGEP-FC 算法流程图

MGEP-FC 算法具体步骤如下：

步骤一：设置 MGEP-FC 初始参数，包括基因头长、DC 域长度、实数集大小、种群规模、锦标赛规模和最大迭代次数；

步骤二：根据具体优化函数设计函数符集，对个体进行编码，生成初始种群并根据具体优化函数计算个体的适应度值；

步骤三：判断是否符合终止条件，不符则转至步骤四，符合则输出种群的最优个体，得到函数最优解；

步骤四：使用式(2)或式(3)衡量种群多样性；

步骤五：判断迭代次数是否大于最大迭代次数/2，满足则进入步骤六，否则进入步骤十；

步骤六：通过模糊控制器更新算法实数集变异率；

步骤七：根据实数集变异率判断实数集是否进行变异操作，是则进入步骤八，否则进入步骤十；

步骤八：重新生成实数集；

步骤九：根据优化函数计算并更新种群个体的适应度值；

步骤十：通过模糊控制器调整算法变异率和交叉率；

步骤十一：进行遗传操作，分别对父代种群进行只交叉、只变异、先交叉后变异和先变异后交叉操作，计算和更新参加遗传操作的个体适应度值；

步骤十二：父代种群中的全部个体以及遗传操

作生成的新个体组成临时种群；

步骤十三：采用锦标赛和精英保留策略对临时种群进行选择操作。先将临时种群中的个体按适应度值进行排序，然后按比例把适应度值最优的个体直接保留到子代种群，最后依据预先设定的锦标赛规模对临时种群进行锦标赛选择。进入步骤三，种群代数加 1。

## 2 实验与讨论

### 2.1 实验一：函数符集设计的效果对比实验

为了增强算法对优化函数的寻优性能，观察优化函数中所含的算子，并在函数符集中使用与其相同的算子。如对表 4 中函数 $f_7$ 进行寻优时，使用函数符集{+，-，*，/，S，C}，更易生成接近最优解的个体。

表 4 测试函数

| 测试函数 | 取值范围 |
| --- | --- |
| $f_1 = \sum_{i=1}^{D} x_i^2$ | $x_i \in [-5.12, 5.12], F_{\min} = 0$ |
| $f_2(x_1, x_2) = 100(x_1^2 - x_2)^2 + (1-x_1)^2$ | $x_i \in [-2.048, 2.048], F_{\min} = 0$ |
| $f_3(x_1, x_2) = 0.5 + \dfrac{\sin^2\sqrt{x_1^2 + x_2^2} - 0.5}{(1.0 + 0.001(x_1^2 + x_2^2))^2}$ | $x_i \in [-100, 100], F_{\min} = 0$ |
| $f_4(x_1, x_2) = (x_1^2 + x_2^2)^{0.25}[\sin^2(50(x_1^2 + x_2^2)^{0.1}) + 1.0]$ | $x_i \in [-100, 100], F_{\min} = 0$ |
| $f_5(x) = \exp(-0.001x)\cos^2(0.8x)$ | $x \in [0, 18], F_{\min} = 0$ |
| $f_6(x, y) = -x\sin(4x) - 1.1y\sin(2y)$ | $x \in [0, 10], y \in [0, 10]$ |
| $f_7(x) = x\sin(10\pi x) + 1.0$ | $x \in [-1, 2]$ |
| $f_8 = [1 + (x_1 + x_2 + 1)^2 \times (19 - 14x_1 + 3x_1^2 - 14x_2 + 6x_1x_2 + 3x_2^2)]$ $\times [30 + (2x_1 - 3x_2)^2 \times (18 - 32x_1 + 12x_1^2 + 48x_2 - 36x_1x_2 + 27x_2^2)]$ | $-2 \leq x_i \leq 2 (i=1,2), F_{\min} = 0$ |
| $f_9 = \sum_{i=1}^{D} i x_i^2$ | $x_i \in [-10, 10], F_{\min} = 0$ |
| $f_{10} = \sum_{i=1}^{D} |x_i| + \prod_{i=1}^{D} |x_i|$ | $x_i \in [-10, 10], F_{\min} = 0$ |
| $f_{11} = \sum_{i=1}^{D} |x_i \sin(x_i) + 0.1 x_i|$ | $x_i \in [-10, 10], F_{\min} = 0$ |
| $f_{12} = -a\exp\left(-b\sqrt{\dfrac{1}{D}\sum_{i=1}^{D} x_i^2}\right) - \exp\left(\dfrac{1}{D}\sum_{i=1}^{D}\cos(c x_i)\right) + a + \exp(1)$ | $a = 20, b = 0.2, c = 2\pi$ $x_i \in [-5.12, 5.12]$ |

设计实验将 MGEP-FC 与使用传统函数符集{+，-，*，/}的 GEP 作对比，除函数符集外其它参数均保持一致。两种方法均随机生成规模为 100 的初始种群，种群分布对比如图 7 所示。在图 7 的(a)(b)中，x 轴为随机生成的函数参数值，y 轴为参数对应的函数值；图(c)(d)为种群个体与其映射的函数值构成的点的分布情况。(a)(c)图为使用特殊函数符集的 MGEP-FC 生成的初始种群，(b)(d)图为使用传统函数符集的 GEP 生成的初始种群。从点的分布情况分析，(a)(c)图明显比(b)(d)图更加均匀，从个体的适应度值分析，(a)(c)图中的个体明显比(b)(c)更优。

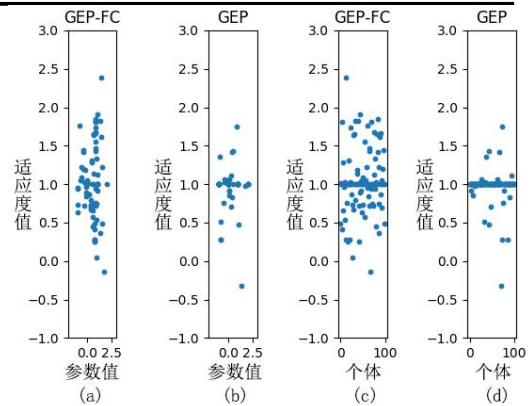

图 7 初始种群个体分布对比图

## 2.2 实验二：遗传操作改进效果对比实验

为验证 1.1.4 节的遗传操作改进效果，选用表 4 的 $f_7$ 函数进行实验。生成规模为 100 的初始种群，个体适应度值分布如图 8 的(a)图所示。

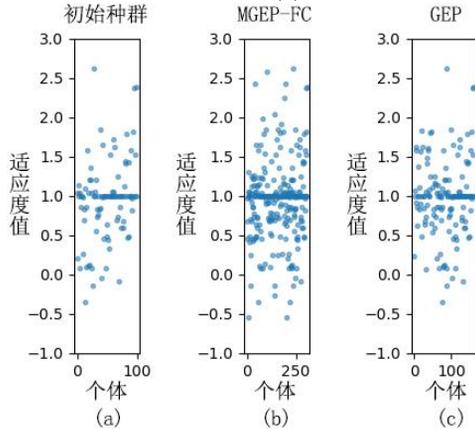

图 8 遗传操作后临时种群个体分布情况对比图

对生成的初始种群使用 1.1.4 节的方法对种群进行遗传操作，得到的临时种群个体适应度值分布如图 8 中(b)图所示。而使用传统 GEP 先交叉后变异方式得到的临时种群，个体适应度值分布如图 8 中(c)图。从图 8 不难看出，同样的初始种群，通过改进后的遗传操作得到的临时种群多样性更优、更有利于优秀基因的保留和进化。

## 2.3 实验三：MGEP-FC 寻优能力测试

本文用 12 个 Benchmark 函数(表 4 所示，其中包括 5 个高维函数，D 表示维数)对 MGEP-FC 算法的寻优能力进行验证。具体算法参数如表 5。实验分为两个部分：第一部分，MGEP-FC 算法与多细胞 GEP 算法对比实验；第二部分，MGEP-FC 算法与其他函数优化算法对比实验。

表 5 算法参数

|  | GEP-MCFO | MGEP-FC |
| --- | --- | --- |
| 函数符集 | {+,-,*,/} | {+,-,*,/,目标函数专用运算符} |
| 锦标赛规模 | 4 | 0.01*临时种群大小 |
| 变异率 | 0.2 | 自适应调整 |
| 交叉率 | 0.3 | 自适应调整 |
| 实数集变异率 | 0.01 | 自适应调整 |
| 普通基因头部长度 |  | 6 |
| 同源基因头部长度 |  | 4 |
| 种群大小 |  | 100 |

### 2.3.1 MGEP-FC 与多细胞 GEP 对比实验

选用函数 $f_1$-$f_8$ 进行实验，其中 MGEP-FC 的交叉率、变异率和实数集变异率使用模糊控制进行动态调整，算法其他参数详见文献[14]。同源基因个数等同于函数的变量数。实验结果如表 6、表 7。对于函数 $f_1$ 和 $f_2$，平均进化代数仅为 1.0638 和 1.46，而函数 $f_3$、$f_4$、$f_5$ 的最优值均产生于初始种群，得益于在函数符集中加入了目标函数专用运算符，使得初始种群就能产生优秀个体。在函数 $f_6$ 中，MGEP-FC 的最优值、历史平均最优和最优值产生代数均有明显提高。对于函数 $f_7$，虽然两种方法在最优值一项中表现相当，但 MGEP-FC 历史平均最优值更优，算法稳定性更高。

表 7 实验结果二

| 项目名 | $f_6$ | | $f_7$ | |
| --- | --- | --- | --- | --- |
|  | GEP-MCFO[14] | MGEP-FC | GEP-MCFO[14] | MGEP-FC |
| 最优函数值 | 18.5547210428 | 18.5547210767 | 2.8502737668 | 2.8502737668 |
| 历史最优平均 | 18.4811463854 | 18.5490834341 | 2.8455127548 | 2.8502732173 |
| 最优值产生于 | 第 8 次的 732 代 | 第 5 次的 447 代 | 第 7 次的 40 代 | 第 1 次的 139 代 |

表 6 实验结果一

| 项目名 | $f_1$ | | $f_2$ | | $f_3$ | | $f_4$ | | $f_5$ | |
| --- | --- | --- | --- | --- | --- | --- | --- | --- | --- | --- |
|  | GEP-MCFO[14] | MGEP-FC | GEP-MCFO[14] | MGEP-FC | GEP-MCFO[14] | MGEP-FC | GEP-MCFO[14] | MGEP-FC | GEP-MCFO[14] | MGEP-FC |
| 最优值 | 0 | 0 | 0 | 0 | 0 | 0 | 0 | 0 | 0 | 0 |
| 成功率 | 100% | 100% | 100% | 100% | 100% | 100% | 100% | 100% | 100% | 100% |
| 产生最优值的平均代数 | 117.85 | 1.0638 | 93.88 | 1.46 | 1.20 | 0 | 257.05 | 0 | 1.00 | 0 |

图 9 为传统多细胞 GEP 算法和 MGEP-FC 算法在迭代过程中种群多样性的变化情况。(a)(c)图分别为两算法迭代过程中临时种群的多样性变化，(b)(d)图分别为其子代种群的多样性波动图。从图 9 明显看出，MGEP-FC 算法生成的临时种群多样性一直保持在较优的状态，而产生的子代种群多样性曲线波动较大，特别是因为实数集的两次变异产生的两次明显波动，表明算法不易陷入局部最优，具有更强的全局搜索能力。

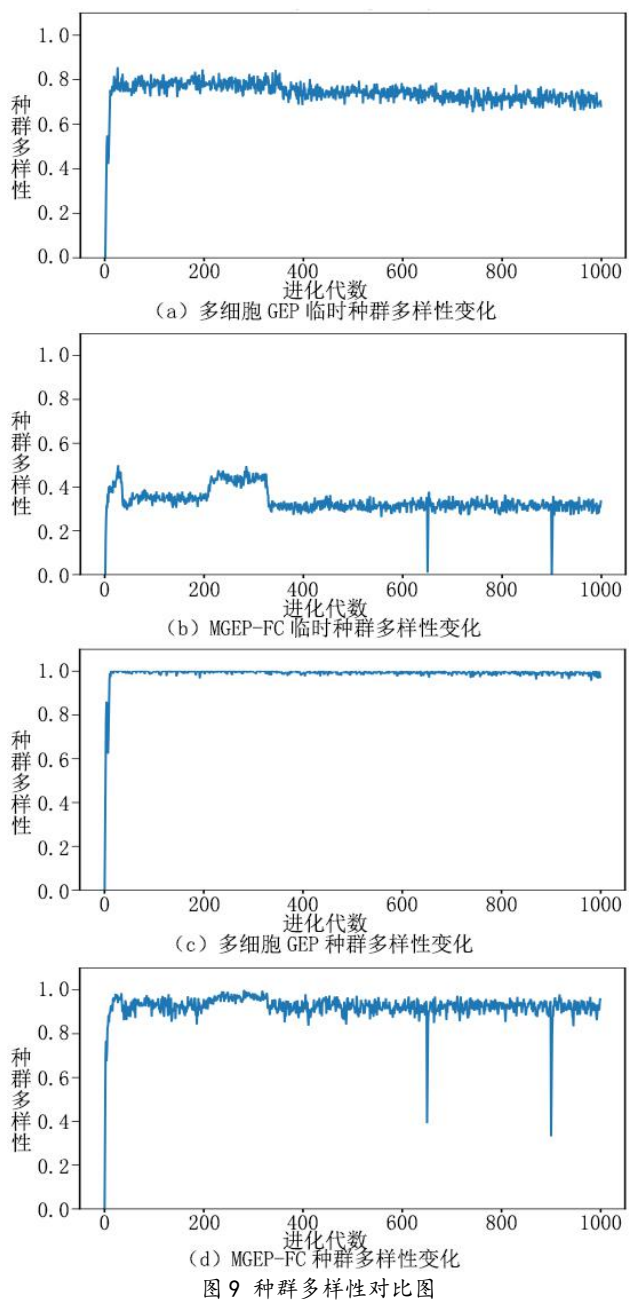

图 9 种群多样性对比图

由于篇幅限制，只展示函数 $f_8$ 的收敛对比图，如图 10 所示。显然，MGEP-FC 相较于 GEP-MCFO 收敛速度明显提高，且解的质量更优。

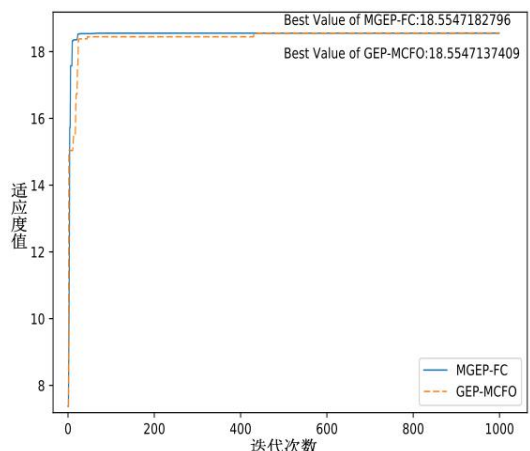

图 10 收敛效果对比图

### 2.3.2 MGEP-FC 与其他函数优化算法对比实验

为了进一步论证本文提出算法的寻优性能，将 MGEP-FC 算法与文献[2]、文献[6]和文献[20]中的 ISABC、RSOS、IABC 等算法进行实验比较，结果如表 8 所示。在函数 $f_8$ 的优化实验中，MGEP-FC 算法能 100%找到函数最优解，并且收敛代数远优于其他两种算法。函数 $f_9$-$f_{12}$ 为高维函数，MGEP-FC 算法的最优解和平均值均比其余算法更优。虽然在函数 $f_{12}$ 中未能找到最优极值，但从均值分析，该算法寻优具有一定的稳定性。

表 8 MGEP-FC 与其他算法对比实验结果

| 优化函数 | 算法 | 运行次数 | 最优值 | 平均值 | 方差 | 平均收敛代数 |
|---|---|---|---|---|---|---|
| $f_8$ | ABC[20] | 50 | 3.0 | 3.0000726 | 7.01E-09 | 336 |
| | IABC[20] | 50 | 3.0 | 3.0000428 | 1.20E-09 | 253 |
| | MGEP-FC | 50 | 3.0 | 3.0 | 0 | 12.38 |
| $f_9$ D=30 | SOS[6] | 30 | 2.49E-122 | 1.87E-116 | 8.65E-116 | - |
| | RSOS[6] | 30 | 0.00E+00 | 1.55E-257 | 0 | - |
| | MGEP-FC | 50 | 0.00E+00 | 0.00E+00 | 0 | 22.84 |
| $f_9$ D=50 | ABC[2] | 30 | 2.48E-15 | 4.72E-15 | 3.87E-30 | - |
| | PSO[2] | 30 | 5.11E-02 | 9.11E-01 | 3.54E-21 | - |
| | ACO[2] | 30 | 8.21E-01 | 9.51E-01 | 8.45E-19 | - |
| | ISABC[2] | 30 | 1.72E-18 | 5.25E-18 | 3.25E-35 | - |
| | MGEP-FC | 50 | 0.00E+00 | 0.00E+00 | 0 | 45.94 |
| $f_{10}$ D=30 | SOS[6] | 30 | 1.22e-61 | 1.18e-57 | 2.67E-57 | - |
| | RSOS[6] | 30 | 1.17e-159 | 7.69e-127 | 2.51e-126 | - |
| | MGEP-FC | 50 | 0.00E+00 | 0.00E+00 | 0 | 22 |
| $f_{10}$ D=50 | ABC[2] | 30 | 9.47E-13 | 2.22E-12 | 1.45E-24 | - |
| | PSO[2] | 30 | 2.65E-01 | 8.21E-01 | 1.33E-29 | - |
| | ACO[2] | 30 | 8.46E-10 | 7.14E-09 | 9.56E-28 | - |
| | ISABC[2] | 30 | 5.14E-16 | 5.98E-16 | 9.22E0-33 | - |
| | MGEP-FC | 50 | 0.00E+00 | 0.00E+00 | 0 | 33.26 |
| $f_{11}$(D=50) | ABC[2] | 30 | 1.68E-05 | 5.02E-05 | 1.08E-09 | - |
| | PSO[2] | 30 | 3.33E-03 | 5.96E-02 | 8.69E-21 | - |
| | ACO[2] | 30 | 6.34E-04 | 5.14E-03 | 5.88E-28 | - |
| | ISABC[2] | 30 | 3.28E-16 | 4.05E-16 | 1.09E-32 | - |
| | MGEP-FC | 50 | 0.00E+00 | 0.00E+00 | 0 | 144.42 |
| $f_{12}$(D=50) | ABC[2] | 30 | 1.35 | 2.36 | 5.81E-01 | - |
| | PSO[2] | 30 | 5.74E+00 | 8.11E+00 | 3.54E-02 | - |
| | ACO[2] | 30 | 6.54E-02 | 6.87E-01 | 4.51E-26 | - |
| | ISABC[2] | 30 | 7.66E-16 | 4.35E-15 | 6.31E-29 | - |
| | MGEP-FC | 50 | 4.44E-16 | 4.44E-16 | 0 | - |

## 3 结束语

针对传统多细胞 GEP 算法在收敛过程中种群多样性易流失、迭代后期易陷入局部最优等问题，提出了一种基于模糊控制的自适应多细胞 GEP 算法（MGEP-FC）。该算法通过衡量每代种群的多样性，结合模糊控制智能调节算法的交叉率、变异率和实数集变异率，使得算法在局部收敛和全局寻优间达到平衡。另外，在遗传操作方式上采用只交叉、只变异、先交叉后变异、先变异后交叉四种方式相结合，生成的新个体和父代种群个体构成临时种群，最后通过锦标赛选择策略在临时种群中选择个体形成子代种群。该操作方案不仅有效避免了迭代过程中种群多样性的大量流失，还增强了个体的发散性寻优能力，种群的整体寻优能力更强。12 个 benchmark 函数的寻优实验结果验证了该算法的有效性和优越性。

虽然 MGEP-FC 算法相较于传统的多细胞 GEP 在收敛代数上有所提升，在进行函数寻优时也能得到令人满意的结果，但是算法的编码和解码阶段依旧需要耗费大量的时间，这也是 GEP 算法待优化的重要方向。